\documentclass[
	paper=a4,
	pagesize=auto,
	fontsize=10pt,
]{scrartcl}

\usepackage[utf8]{inputenc}
\usepackage[
	english,
]{babel}

\makeatletter
\newcommand{\LoadPackagesNow}{}
\newcommand{\LoadPackageLater}[2][]{%
	\g@addto@macro{\LoadPackagesNow}{%
		\usepackage[#1]{#2}%
	}%
}
\makeatother

\usepackage{xspace}
\usepackage{xargs}
\usepackage{ifthen}
\usepackage{etoolbox}

\KOMAoptions{%
	numbers=noenddot,
	twoside=semi, 
}

\setkomafont{section}{\Large\rmfamily\bfseries}
\setkomafont{subsection}{\large\rmfamily\bfseries}
\setkomafont{subsubsection}{\rmfamily\bfseries}
\setkomafont{paragraph}{\rmfamily\bfseries}
\setkomafont{pageheadfoot}{\smaller\scshape\rmfamily}

\usepackage[headsepline,automark]{scrlayer-scrpage}
\clearmainofpairofpagestyles
\clearplainofpairofpagestyles
\rehead{\pagemark}
\rohead{\pagemark}
\lehead{\headmark}
\lohead{\shortauthor}
\automark[section]{section}

\usepackage{calc}
\usepackage[margin=3cm,a4paper]{geometry}
\usepackage{framed}
\usepackage{mdframed}
\usepackage{enumitem}
\usepackage[toc]{appendix}

\usepackage[
bottom,
stable,      
multiple,    
flushmargin,
hang,
]{footmisc}
\usepackage{mathpazo} 

\usepackage{microtype}

\usepackage{relsize}
\usepackage[normalem]{ulem}
\usepackage{soul}
\usepackage{url}
\usepackage{lipsum}

\usepackage{array}
\usepackage{multirow}
\usepackage{longtable}
\usepackage{hhline}
\usepackage{tabularx}
\usepackage{booktabs}
\usepackage{setspace}

\usepackage[table]{xcolor}
\usepackage{pgfplots}
\pgfplotsset{compat=1.15}
\usepackage{tikz}
\usepackage{tkz-euclide}
\usetikzlibrary{arrows,arrows.meta,shapes,calc,decorations.pathreplacing,decorations.pathmorphing,
	intersections,quotes,angles,positioning,fit,petri,through,backgrounds}
\tikzstyle{blackdot}=[shape=circle,fill=black,minimum size=1mm,inner sep=0pt,outer sep=0pt]

\usepackage{graphicx}

\usepackage{caption}
\usepackage{subcaption}
\usepackage[
tbtags,    
sumlimits,  
nointlimits, 
namelimits, 
reqno,     
]{amsmath}

\usepackage{amsfonts}
\usepackage{mathrsfs}
\usepackage{dsfont}
\usepackage{amssymb}
\usepackage{units}
\LoadPackageLater{amsthm}
\LoadPackageLater{thmtools}
\usepackage[fixamsmath,disallowspaces]{mathtools}
\mathtoolsset{showonlyrefs}
\mathtoolsset{centercolon=true}

\usepackage{bm} 
\makeatletter
\g@addto@macro\bfseries{\boldmath}
\makeatother

\allowdisplaybreaks[4] 
\numberwithin{equation}{section} 

\usepackage{csquotes}
\usepackage[
style=alphabetic,
sorting=ynt,
sortcites=true,
giveninits=true,
maxbibnames=99,
backend=biber,
maxalphanames=5,
maxcitenames=2,
]{biblatex}
\renewbibmacro{in:}{}
\DeclareFieldFormat{pages}{#1}

\usepackage[textsize=tiny,english,colorinlistoftodos,
	disable,
]{todonotes}

\definecolor{pdfurlcolor}{rgb}{0,0,0.6}
\definecolor{pdffilecolor}{rgb}{0.7,0,0}
\definecolor{pdflinkcolor}{rgb}{0,0,0.6}
\definecolor{pdfcitecolor}{rgb}{0,0,0.6}
\usepackage[
colorlinks=true,
urlcolor=pdfurlcolor,
filecolor=pdffilecolor,
linkcolor=pdflinkcolor,
citecolor=pdfcitecolor,
raiselinks=true,
breaklinks,
verbose,
hyperindex=true,
linktocpage=true,
hyperfootnotes=false,
bookmarks=true,
bookmarksopenlevel=3,
bookmarksopen=true,
bookmarksnumbered=true,
bookmarkstype=toc,
bookmarksdepth=4,
plainpages=false,
pageanchor=true,
pdfstartview=FitH,
pdfpagemode=UseOutlines,
pdfpagelabels=true,
pdfpagelayout=OneColumn,
]{hyperref}

\LoadPackagesNow 

\newcommand{\ifargdef}[3][{}]{\ifthenelse{\equal{#2}{}}{#1}{#3}}

\hypersetup{
	pdfauthor={Martin Genzel, Jan Macdonald, Maximilian März},
	pdftitle={Report on DL-Sparse-View CT Challenge},
	pdfsubject={Article},
	pdfcreator={PDF-LaTeX},
}

\newenvironment{rmklist}
{\begin{enumerate}[label={(\arabic*)},itemindent=2em,leftmargin=0em]}
{\end{enumerate}}

\newtheoremstyle{claim}
	{\topsep}{\topsep}%
	{\itshape}
	{}
	{}
	{}
	{.5em}
	{{\bfseries\boldmath\thmname{#1} \thmnumber{#2}} \thmnote{(#3)}}

\newtheoremstyle{definition}
	{\topsep}{\topsep}%
	{}
	{}
	{}
	{}
	{.5em}
	{\textbf{\thmname{#1} \thmnumber{#2}} \thmnote{(#3)}}
	
\newtheoremstyle{algorithm}
	{\topsep}{\topsep}%
	{}
	{}
	{\bfseries\boldmath}
	{}
	{\newline}
	{\thmname{#1} \thmnumber{#2} \thmnote{(#3)}}

\mdfdefinestyle{boxed}{innertopmargin =6pt, splittopskip = \topskip, skipbelow=6pt, skipabove=12pt}
\mdfdefinestyle{emphframe}{linecolor=black, innertopmargin = 3pt, splittopskip = \topskip, skipbelow=6pt, skipabove=6pt}

\declaretheorem[style=claim,numberwithin=section]{theorem}

\declaretheorem[style=definition,sibling=theorem,qed=$\Diamond$]{remark}

\declaretheorem[style=algorithm,sibling=theorem,%
	preheadhook={\begin{mdframed}[style=emphframe] \setcounter{mpfootnote}{\value{footnote}}},%
	postfoothook=\end{mdframed}]{experiment}
\declaretheorem[style=algorithm,sibling=theorem,%
	preheadhook={\begin{mdframed}[style=emphframe] \setcounter{mpfootnote}{\value{footnote}}},%
	postfoothook=\end{mdframed}]{algorithm}
\declaretheorem[style=definition,sibling=theorem,%
	preheadhook={\begin{mdframed}[style=boxed] \setcounter{mpfootnote}{\value{footnote}}},%
	postfoothook=\end{mdframed}]{recipe}

\newcommand{\opleft}[1]{\mathopen{}\left#1}
\newcommand{\opright}[1]{\right#1\mathclose{}}
\newcommandx{\braces}[4]{%
\ifstrequal{#3}{normal}{#1#4#2}{%
\ifstrequal{#3}{auto}{\left#1#4\right#2}{%
\ifstrequal{#3}{opauto}{\opleft#1#4\opright#2}{%
#3#1#4#3#2}}}%
}

\newcommand{\R}{\mathbb{R}} 

\newcommand{\suchthat}[1][normal]{\ifstrequal{#1}{normal}{\mid}{#1|}} 

\newcommandx{\intvcl}[3][1=normal]{\braces{[}{]}{#1}{#2, #3}} 
\newcommandx{\intvop}[3][1=normal]{\braces{(}{)}{#1}{#2, #3}} 
\newcommandx{\intvclop}[3][1=normal]{\braces{[}{)}{#1}{#2, #3}} 
\newcommandx{\intvopcl}[3][1=normal]{\braces{(}{]}{#1}{#2, #3}} 

\newcommandx{\abs}[2][1=normal]{\braces{\lvert}{\rvert}{#1}{#2}} 
\newcommandx{\ceil}[2][1=normal]{\braces{\lceil}{\rceil}{#1}{#2}} 
\newcommandx{\floor}[2][1=normal]{\braces{\lfloor}{\rfloor}{#1}{#2}} 
\newcommandx{\round}[2][1=normal]{\braces{[}{]}{#1}{#2}} 
\newcommandx{\der}[1]{D^{#1}} 
\newcommandx{\gradient}{\nabla} 
\newcommandx{\partder}[4][1={},4={}]{\frac{\partial^{#4} #2}{\partial #3^{#4}}\ifargdef{#1}{\Big|_{#1}}} 
\newcommandx{\integ}[4][1={},2={}]{\int_{#1}^{#2} #3 \, #4} 
\newcommandx{\asympffaster}[2][1=normal]{o\braces{(}{)}{#1}{#2}} 
\newcommandx{\asympfaster}[2][1=normal]{O\braces{(}{)}{#1}{#2}} 
\newcommandx{\asympeq}[2][1=normal]{\Theta\braces{(}{)}{#1}{#2}} 
\newcommandx{\asympsslower}[2][1=normal]{\omega\braces{(}{)}{#1}{#2}} 
\newcommandx{\asympslower}[2][1=normal]{\Omega\braces{(}{)}{#1}{#2}} 

\newcommandx{\norm}[2][1=normal]{\braces{\|}{\|}{#1}{#2}} 
\renewcommandx{\sp}[3][1=normal]{\braces{\langle}{\rangle}{#1}{#2, #3}} 
\newcommandx{\End}[2][2={}]{\mathcal{L}\opleft( #1 \ifargdef{#2}{, #2} \opright)} 
\newcommand{\T}{\mathsf{T}} 
\renewcommand{\vec}[1]{\boldsymbol{#1}} 
\newcommandx{\opnorm}[2][1=normal]{\norm[#1]{#2}_{\operatorname{op}}} 
\newcommandx{\ball}[2][1={},2={}]{B_{#1}^{#2}} 

\newcommandx{\measure}[2][1=normal]{\operatorname{vol}\braces{(}{)}{#1}{#2}} 
\newcommandx{\Leb}[3][1={},3=normal]{L^{#2}\ifargdef{#1}{\braces{(}{)}{#3}{#1}}{}} 
\newcommandx{\Lebnorm}[4][1=normal,3={2},4={}]{\norm[#1]{#2}_{#3}} 
\renewcommandx{\l}[3][1={},3=normal]{\ell_{#2}\ifargdef{#1}{\braces{(}{)}{#3}{#1}}} 
\newcommandx{\lnorm}[4][1=normal,3={2},4={}]{\norm[#1]{#2}_{#3}} 
\newcommandx{\Smooth}[4][1={},3={},4=normal]{C_{#3}^{#2}\ifargdef{#1}{\braces{(}{)}{#4}{#1}}} 
\newcommandx{\Schwartz}[2][1={},2=normal]{\mathscr{S}\ifargdef{#1}{\braces{(}{)}{#2}{#1}}} 
\newcommandx{\Schwartzpoly}[2][1=normal]{\braces{\langle}{\rangle}{#1}{\abs[#1]{#2}} } 
\newcommandx{\Tempdistr}[2][1={},2=normal]{\mathscr{S}'\ifargdef{#1}{\braces{(}{)}{#2}{#1}}} 
\newcommandx{\distrinp}[3][1=normal]{\braces{\langle}{\rangle}{#1}{#2, #3}} 
\newcommandx{\ft}[3][1=default,2=auto]{
\ifstrequal{#1}{default}{\widehat{#3}}{
\ifstrequal{#1}{long}{{\braces{(}{)}{#2}{#3}}^{\wedge}}{}}} 
\newcommandx{\ift}[3][1=default,2=auto]{
\ifstrequal{#1}{default}{\check{#3}}{
\ifstrequal{#1}{long}{{\braces{(}{)}{#2}{#3}}^{\vee}}{}}} 

\newcommandx{\prob}[2][1={},2=normal]{\mathbb{P}\ifargdef{#1}{\braces{[}{]}{#2}{#1}}}
\newcommandx{\mean}[2][1={},2=normal]{\mathbb{E}\ifargdef{#1}{\braces{[}{]}{#2}{#1}}}
\newcommandx{\var}[2][1={},2=normal]{\mathbb{V}\ifargdef{#1}{\braces{[}{]}{#2}{#1}}}

\newcommandx{\Unif}[2][1=normal]{\mathcal{U}\braces{(}{)}{#1}{#2}} 
\newcommandx{\Normdistr}[3][1=normal]{\mathcal{N}\braces{(}{)}{#1}{#2, #3}} 
\newcommandx{\Poi}[2][1=normal]{\mathrm{Poi}\braces{(}{)}{#1}{#2}} 
\newcommandx{\normsubg}[2][1=normal]{\norm[#1]{#2}_{\psi_2}} 

\newcommand{\y}{\vec{y}} 
\newcommand{\ygrtr}{\vec{y}_0} 

\newcommand{\x}{\vec{x}} 
\newcommand{\xgrtr}{\vec{x}_0} 

\newcommand{\DC}{\mathcal{DC}}
\newcommand{\dcparam}{\lambda}
\newcommand{\DCparam}{\vec{\lambda}}
\newcommand{\NNparam}{\vec{\theta}}
\newcommand{\Fanparam}{\vec{\theta}_{\smash{\texttt{fan}}}}
\newcommand{\AFan}{\vec{F}} 
\newcommand{\FBP}{\texttt{FBP}} 

\newcommand{\UNet}{\textrm{UNet}}

\newcommand{\Tira}{\textrm{Tiramisu}}

\newcommand{\ItNet}{\textrm{ItNet}}
\newcommand{\ItNett}{\textrm{ItNet-post}}
\newcommand{\U}{\vec{U}}

\begin{document}
\newcommand{\ToDo}[1]{{\color{red}#1}}

\pagestyle{scrheadings}

\renewcommand*{\thefootnote}{\fnsymbol{footnote}}

\begin{center}
{\bfseries\larger[2]{{\smaller AAPM DL-Sparse-View CT Challenge Submission Report:}\\[.5ex] Designing an Iterative Network for Fanbeam-CT \\ with Unknown Geometry}}
\end{center}

\vspace{0.5\baselineskip}
\begin{addmargin}[2em]{2em}

\begin{center}
\noindent{\normalsize{\textbf{Martin Genzel}\footnote{Mathematical Institute, Utrecht University, Netherlands} \qquad \textbf{Jan Macdonald}\footnote{Institute of Mathematics, Technische Universität Berlin, Germany} \qquad \textbf{Maximilian März}\footnotemark[2]}}
\end{center}

\vspace{1\baselineskip}
{\smaller
\noindent\textbf{Abstract.}
This report is dedicated to a short motivation and description of our contribution to the AAPM DL-Sparse-View CT Challenge (team name: \texttt{robust-and-stable}).
The task is to recover breast model phantom images from limited view fanbeam measurements using data-driven reconstruction techniques.
The challenge is distinctive in the sense that participants are provided with a collection of ground truth images and their noiseless, subsampled sinograms (as well as the associated limited view filtered backprojection images), but not with the actual forward model.
Therefore, our approach first estimates the fanbeam geometry in a data-driven geometric calibration step.
In a subsequent two-step procedure, we design an iterative end-to-end network that enables the computation of near-exact solutions.

}

\vspace{.1\baselineskip}

\end{addmargin}
\newcommand{\shortauthor}{Genzel, Macdonald, März: DL-Sparse-View CT Challenge}

\renewcommand*{\thefootnote}{\arabic{footnote}}
\setcounter{footnote}{0}

\thispagestyle{plain}

\begin{mdframed}[outerlinewidth=2,innerlinewidth=0,leftmargin=40,rightmargin=40,outerlinecolor=gray,roundcorner=2]
	\smaller
	\textbf{Note:} This is a technical report of a method participating in a \emph{not yet finished} challenge. Therefore, it does not contain any final results. In particular, the reported reconstruction errors are only with respect to our own validation split of the provided training data. Once the official challenge report is released, these values will be updated with the results from the actual test set.
\end{mdframed}

\section{Introduction}

\enlargethispage{1\baselineskip}
In recent years, deep learning methods have been successfully applied to problems of the natural sciences.
A prominent example of such \emph{scientific machine learning} is the development of efficient solutions strategies for inverse problems, such as those encountered in the context of medical imaging.
Despite unprecedented empirical performance in various practical scenarios, a sound theoretical understanding of data-driven reconstruction methods seems to be out of reach to date.

For this reason, more and more critical voices are heard, questioning the reliability of deep-learning-based solution strategies.
For instance, Sidky et al.~\cite{sidky20} have recently demonstrated that \emph{post-processing} by the prominent $\UNet$-architecture may not yield satisfactory recovery precision in a sparse-view computed tomography (CT) scenario.
These findings have led to the AAPM Challenge, to which the present report is devoted.
The goal of the challenge is \emph{``to identify the state-of-the-art in solving\footnote{This expression is used to describe methods that provide perfect recovery in the idealistic situation of noiseless measurements. As we understand it, the greater goal of the AAPM challenge is to evaluate whether it is also possible to achieve such a (near-)exact precision by deep-learning-based schemes.} the CT inverse problem with data-driven techniques''} \cite{sid+21}.

A different and much-noticed example of such a critical perspective is the work~\cite{arpah20}, which claims that \emph{``deep learning typically yields unstable methods for image reconstruction''}.
Addressing this concern, we have recently examined a representative selection of end-to-end networks in the context of inverse problems~\cite{genzel20}.
Surprisingly and in contrast to~\cite{arpah20}, our study has demonstrated that deep-learning-based recovery schemes are very stable to measurement perturbations.
It goes without saying that the ability to \emph{accurately} solve an inverse problem also plays a central role in that regard.
To cover this aspect, \cite{genzel20} has considered scenarios that are similar to the sparse-view setup of the AAPM challenge in the following sense:
\emph{Exact} signal recovery by classical total variation (TV) minimization is possible for \emph{noisefree} measurements.
In such situations, we were also able to train neural networks (NNs) that provide visually perfect reconstructions.
We took this as a motivation to participate in the AAPM challenge with the goal of designing a data-driven recovery workflow for (near-)exact image recovery.

\paragraph{Our Strategy in a Nutshell.}

Our approach is rooted in the following (debatable) observation:

\begin{mdframed}[outerlinewidth=2,innerlinewidth=0,leftmargin=40,rightmargin=40,backgroundcolor=gray!20,outerlinecolor=gray,roundcorner=2]
 High reconstruction accuracy is only possible when the forward model is explicitly incorporated into the reconstruction mapping, e.g., by an iterative promotion of data-consistent solutions.
\end{mdframed}

\newpage
Highlighting the importance of incorporating the forward operator is by no means novel.
It is one of the central pillars of scientific machine learning, where NNs are frequently enriched (or constrained) by physical modeling.
Indeed, the seminal works on deep-learning-based solution strategies for inverse problems are inspired by unrolling classical algorithms~\cite{kl10,yslx16}.
Furthermore, current state-of-the-art methods seem to be exclusively based on \emph{iterative} end-to-end networks, e.g., see~\cite{kno+20b,muc+20,leuschner21}.

Our approach to the AAPM challenge and its contributions in a broader sense can be summarized as follows:

\begin{enumerate}[label={(\roman*)}]
 \item Given that the exact forward model is unknown, we pursue a data-driven estimation of the underlying fanbeam geometry.
 This is achieved by fitting a generic, parameterized fanbeam operator to the provided sinogram-image pairs in a deep-learning-like fashion (i.e., by gradient descent with backpropagation/automatic differentiation).
 We hope that this approach can be of further use in the context of geometric calibration and forward operator correction.
 In particular, we currently explore an unsupervised identification strategy based on sinogram consistency conditions.

 \item We propose a conceptually simple, yet powerful deep-learning workflow, which turns a post-processing UNet~\cite{rfb15} into an iterative reconstruction scheme.
 From a technical perspective, most of its design components have been previously reported in the literature.
 However, the overall strategy appears to be novel and differs from more common unrolled networks in several aspects, including:
 (a)~we make use of a \emph{pre-trained} UNet as the computational backbone;
 (b)~data-consistency is inspired by an $\ell^2$-gradient step, but utilizes the \emph{filtered} backprojection instead of the regular adjoint.
 We think that the proposed strategy will be of use for other inverse problems as well, given that it outperforms other state-of-the-art data-driven approaches, such as the learned primal dual algorithm~\cite{ao18}, by an order of magnitude with respect to the root-mean-square-error (RMSE).

\end{enumerate}

\section{Methodology}

In this section, we give a short overview of our approach, together with a motivation of some design choices.
A public code repository can be found under \cite{gmm21git}.

\paragraph{Step 1 -- Data-Driven Geometry Identification.}

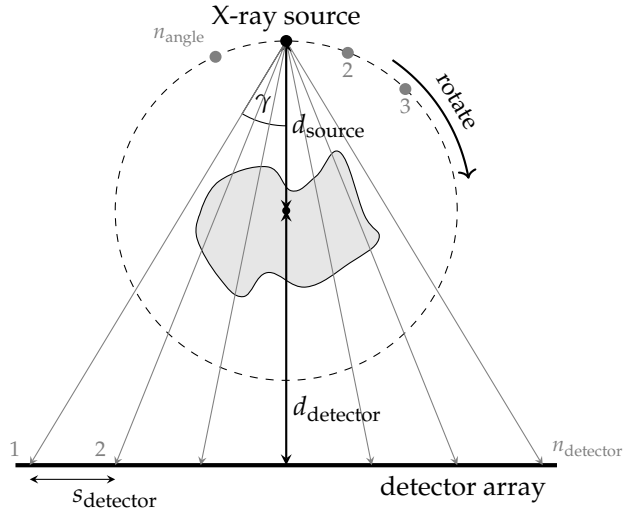
\begin{figure}
	\centering
	\begin{tikzpicture}[scale=0.75]

\draw[dashed] (0.5, 0.5) circle[radius=3.0];

\draw[black, ultra thick] (-4.25,-4) -- (5.25,-4) node[below left]{detector array};

\draw[stealth-stealth, black] (-4,-4.25) -- (-2.5,-4.25) node[below]{$s_{\text{detector}}$};

\draw[<-,domain=10:55,thick] plot ({0.5+3.25*cos(\x)}, {0.5+3.25*sin(\x)});
\draw (3.25,2.3) node[rotate=-60,above,scale=0.9,thick] {rotate};

\draw[fill=black, fill opacity=0.1,scale=0.85]  plot[smooth, tension=.7] coordinates {(-1.2,0.5) (-0.8,1.2) (0.2,1.5) (0.7,1) (1.2,1.3) (1.7,1.8) (2.2,0.5) (2.5, 0) (1.9,-0.5) (1.0,-1) (0.2,-0.7) (-0.3,-1.2) (-0.8,-0.8) (-1.25, 0) (-1.2,0.5)};

\draw (0.5,3.5) -- ++(-90:1.5) arc (-90:-121:1.5) node[midway,above]{$\gamma$} -- cycle;


\draw[-stealth,gray] (0.5,3.5) -- (5,-4) node[above right]{\footnotesize $n_{\text{detector}}$};

\draw[-stealth,gray] (0.5,3.5) -- (3.5,-4);

\draw[-stealth,gray] (0.5,3.5) -- (2,-4);

\draw[thick, stealth-stealth, black] (0.5,3.5) -- (0.5,0.5);
\draw (1.25,2) node[black] {$d_{\text{source}}$};

\draw[thick, stealth-stealth, black] (0.5,0.5) -- (0.5,-4);
\draw (1.4,-3) node[black] {$d_{\text{detector}}$};

\draw[-stealth,gray] (0.5,3.5) -- (-1.0,-4); 

\draw[-stealth,gray] (0.5,3.5) -- (-2.5,-4) node[above left]{\footnotesize 2};

\draw[-stealth,gray] (0.5,3.5) -- (-4.0,-4) node[above left]{\footnotesize 1};

\fill[thick, black] (0.5,0.5) circle (2pt);

\fill[thick, black] (0.5,3.5) circle (3pt) node[above] {X-ray source};
\fill[thick, gray] (1.58,3.29) circle (3pt) node[below]{\footnotesize 2};
\fill[thick, gray] (2.59,2.65) circle (3pt) node[below]{\footnotesize 3};
\fill[thick, gray] (-0.74,3.22) circle (3pt) node[above left]{\footnotesize $n_\text{angle}$};
\end{tikzpicture}
	\caption{\textbf{Fanbeam geometry.} Illustration of the parameters determining the geometry of the fanbeam CT model.}
	\label{fig:radon}
\end{figure}

The first step of our reconstruction pipeline learns the unknown forward operator from the provided training data.
The continuous version of \emph{tomographic fanbeam measurements} is based on computing line integrals:
\begin{equation*}
    p(s,\varphi) = \int_{L(s,\varphi)} x_0(x,y) \, \mathrm{d}(x,y),
\end{equation*}
where $x_0$ is the unknown image and $L(s,\varphi)$ denotes a line in fanbeam coordinates, i.e., $\varphi$ is the \emph{fan rotation angle} and $s$ encodes the \emph{sensor position}; see \cite{fessler17} for more details.
In an idealized\footnote{We have found that this basic model was enough to accurately describe the AAPM challenge setup. If needed, it would be possible to account for other factors such as non-flat detector arrays, offsets of the axis of rotation from the origin, misalignments of the detector array, etc.} situation, the fanbeam model is specified by the following geometric parameters (see Fig.~\ref{fig:radon}):
\begin{itemize}
    \item $d_\text{source}$ -- the distance of the X-ray source to the origin,
    \item $d_\text{detector}$ -- the distance of the detector array to the origin,
    \item $n_\text{detector}$ -- the number of detector elements,
    \item $s_\text{detector}$ -- the spacing of the detector elements along the array,
    \item $n_\text{angle}$ -- the number of fan rotation angles,
    \item $\vec{\varphi}\in [0,2\pi]^{n_\text{angle}}$ -- a discrete list  of rotation angles.
\end{itemize}
Here, it is assumed that integrals are only measured along a finite number of lines, determined by $m \coloneqq n_\text{detector} \cdot n_\text{angle}$.
In the AAPM DL-\emph{Sparse-View} Challenge, the resulting forward operator is \emph{severely ill-posed}, since only the measurements of a few fan rotation angles $n_\text{angle}$ are acquired.
Furthermore, the geometric setup is not disclosed to the challenge participants --- it is only known that fanbeam measurements are taken.

We have addressed this lack of information by a data-driven estimation strategy that fits the above set of parameters to the given training data.
To this end, we first observe that the previous parametrization is redundant, and without of loss of generality, we may assume that $s_\text{detector}=1$ (by rescaling $d_\text{detector}$ appropriately).
Further, if the field-of-view angle $\gamma$ is known, then the relation
\begin{equation}
\label{eq:simp1}
d_\text{detector} = \frac{n_\text{detector} \cdot s_\text{detector}}{2\tan{\gamma}}-d_\text{source}
\end{equation}
can be used to eliminate another parameter.
Thus, the fanbeam geometry is effectively determined by the reduced parameter set $(d_\text{source}, n_\text{detector}, n_\text{angle}, \vec{\varphi})$.
The training data provides pairs of discrete images $\xgrtr \in\R^{\smash{512\cdot 512 \eqqcolon N}}$ and its simulated fanbeam measurements $\ygrtr \in\R^{\smash{128\cdot1024 = m}}$, from which the dimensions $n_{\smash{\text{angle}}}=128$ and $n_{\smash{\text{detector}}} = 1024$ can be derived.
We determine the field of view as $\gamma = \arcsin(256 / d_\text{source})$, so that the  maximum inscribed circle in the discrete image is exactly contained within each fan of lines, which is a common choice for fanbeam CT.
Hence, \eqref{eq:simp1} leads to
\begin{equation*}
d_\text{detector} = 2 \cdot s_\text{detector} \cdot \sqrt{d_\text{source}^2  - 256^2} - d_\text{source} \ .
\end{equation*}

The \textbf{main difficulty of Step 1} lies in the estimation of the remaining parameters $(d_\text{source},\vec{\varphi})$.
To that end, we have implemented a discrete fanbeam transform from scratch in \texttt{PyTorch} (together with its corresponding filtered backprojection). 
A distinctive aspect of our implementation is the use of a vectorized numerical integration that enables the efficient computation of derivatives with respect to the geometric parameters by means of \emph{automatic differentiation}.
This feature can be exploited for a data-driven parameter identification, for instance, by a gradient descent.
More precisely, we use a ray-driven numerical integration for the forward model and a pixel-driven and sinogram-reweighting-based filtered backprojection (with a Hamming filter) \cite[Sec. 3.9.2]{fessler17}.
In addition to the parameters $(d_\text{source},\vec{\varphi})$, we also introduce learnable scaling factors $s_\text{fwd}$ and $s_\text{fbp}$ for the forward and inverse transform, respectively.
They account for ambiguities in chosing the discretization units of distance compared to the actual physical units of distance.

As previously indicated, we estimate the free parameters $\Fanparam = (s_{\smash{\text{fwd}}},d_{\smash{\text{source}}},\vec{\varphi})\in\R^{130}$ of the implemented forward operator $\AFan[\Fanparam]\in\R^{m\times N}$ in a deep-learning-like fashion:
The ability to compute derivatives $\tfrac{\mathrm{d}\AFan}{\mathrm{d}\Fanparam}$ allows us to make use of the $M=4000$ sinogram-image pairs $\{(\ygrtr^i, \xgrtr^i) \}_{i=1}^M$ by solving
\begin{equation}\label{eq:op_ident}
  \min_{\Fanparam} \ \tfrac{1}{M}\sum_{i=1}^M \norm[\big]{\AFan[\Fanparam] (\xgrtr^i) - \ygrtr^i}_2^2 
\end{equation}
with a variant of gradient descent (see Remark~\ref{rmk:rmk1} for details).
Finally, we determine $s_\text{fbp}$ by solving
\begin{equation}\label{eq:fpb_ident}
  \min_{s_\text{fbp}} \ \tfrac{1}{M}\sum_{i=1}^M \norm[\big]{\xgrtr^i - \FBP[\Fanparam,s_\text{fbp}](\ygrtr^i)}_2^2 \ , 
\end{equation}
while keeping the already identified parameters fixed.
From now on, we will use the short-hand notation $\AFan$ and $\FBP$ for the estimated operators $\AFan[\Fanparam]$ and $\FBP[\Fanparam,s_\text{fbp}]$, respectively.

\begin{remark}
\label{rmk:rmk1}
\begin{rmklist}

    \item Clearly, the formulation~\eqref{eq:op_ident} is non-convex and therefore it is not clear whether gradient descent enables an accurate estimation of the underlying fanbeam geometry.
    Indeed, standard gradient descent was found to be very sensitive to the initialization of $\Fanparam$ and got stuck in bad local minima.
    To overcome this issue, we solve~\eqref{eq:op_ident} by a \emph{coordinate descent} instead, which alternatingly optimizes over $s_\text{fwd}$, $d_\text{source}$, and $\vec{\varphi}$ with individual learning rates.
    This strategy was found to effectively account for large deviations of gradient magnitudes of the different parameters.
    Indeed, we observed a fast convergence and a reliable identification of $\Fanparam$, independently of the initialization.

    \item In principle, the strategy of~\eqref{eq:op_ident} requires only few training samples to be successful.
    However, when verifying the robustness of the outlined strategy against measurement noise, we observed that it is beneficial to employ more training data.

    \item Subsequent to the estimation of an accurate fanbeam geometry, we still noted a systematic error in our forward model.
    We suspect that it is caused by subtle differences in the numerical integration in comparison to the true forward model of the AAPM challenge.
    In compensation, we compute the (pixelwise) mean error over the training set, as an additive correction of the model bias.

 \end{rmklist}
\end{remark}

\paragraph{Step 2 -- Pre-Training a UNet as Computational Backbone.}

The centerpiece of our reconstruction scheme is formed by a standard UNet-architecture $\U[\NNparam] \colon \R^N \to \R^N$~\cite{rfb15}.
It is first employed as a residual network to post-process sparse-view filtered backprojection images, i.e., we consider the reconstruction mapping
\begin{equation}
 \UNet[\NNparam] \colon \R^m \to \R^N, \ \y \mapsto \left[\U[\NNparam] \circ \FBP\right] (\y).
\end{equation}
The learnable parameters $\NNparam$ are trained from the collection of $M=4000$ sinogram-image pairs $\{(\ygrtr^i, \xgrtr^i) \}_{i=1}^M$ that are provided by the challenge.
This is achieved by standard empirical risk minimization, i.e., by (approximately) solving
\begin{equation}
\label{eq:erm}
 \min_{\NNparam} \ \tfrac{1}{M}\sum_{i=1}^M \norm[\big]{\xgrtr^i - \UNet[\NNparam] (\ygrtr^i)}_2^2 + \mu \cdot \norm{\NNparam}_2^2 \ ,
\end{equation}
where we choose $\mu = 10^{-3}$.
This minimization problem is tackled by $400$ epochs of mini-batch stochastic gradient descent and the Adam optimizer~\cite{kb14} with initial learning rate $0.0002$ and batch size $4$.

\enlargethispage{2\baselineskip}
\begin{remark}
The post-processing strategy of Step 2 was pioneered in~\cite{kmy17,che+17b} and popularized by~\cite{jmfu17,che+17}, among many others.
Due to the multi-scale encoder-decoder structure with skip-connections, the $\UNet$-architecture is very efficient in handling image-to-image problems.
Therefore, solving~\eqref{eq:erm} typically works out-of-the-box without requiring sophisticated initialization or optimization strategies (even in seemingly hopeless situations~\cite{hauptmann20}).
Making use of a more powerful or a more memory-efficient network would be beneficial, e.g., see results for the Tiramisu network below.
However, we preferred to keep our workflow as simple as possible and therefore decided to stick to the standard $\UNet$ as the main computational building block.
\end{remark}

\paragraph{Step 3 -- Constructing an Iterative Scheme.}

In this step, we discuss our main reconstruction method.
It incorporates the (approximate) forward model $\AFan$ from Step 1 (and the associated inversion by the $\FBP$) via the following iterative procedure:
\begin{align}
 \label{eq:itnet}
 \ItNet[\NNparam] \colon \R^m \to \R^N, \ \y \mapsto \left[ \bigcirc_{k=1}^4\left( \DC_{\dcparam_k,\y} \circ \U[\tilde{\NNparam}] \right) \circ \FBP \right](\y),
\end{align}
for the learnable parameters $\NNparam = [\tilde{\NNparam},\lambda_1,\lambda_2,\lambda_3,\lambda_4]$ and the $k$-th \emph{data-consistency} layer
\begin{align}
 \DC_{\dcparam_k,\y} \colon \R^N \to \R^N, \ \x \mapsto \x - \dcparam_k \cdot \FBP (\AFan \x - \y).
\end{align}
$\ItNet$ is trained by empirical risk minimization analogously to~\eqref{eq:erm} with $\mu=10^{-4}$.
We run $500$ epochs of mini-batch stochastic gradient descent and Adam with an initial learning rate of $8\cdot 10^{-5}$ and a batch size of $2$ (restarting Adam after $250$ epochs).
The $\UNet$-parameters $\tilde{\NNparam}$ are initialized by the weights obtained in Step 2.

In the following, we will briefly discuss central aspects of the architecture in~\eqref{eq:itnet} and motivate some of the important design choices:
\begin{enumerate}[label={(\roman*)}]
 \item The computational centerpiece of $\ItNet$ is formed by the \textbf{$\UNet$}-architecture.
 This stands in contrast to earlier generations of unrolled iterative schemes, which rely on basic convolutional blocks instead, e.g., see~\cite{ao18,yslx16}.
 We have found that it is advantageous to exploit the efficacy of $\UNet$-like image-to-image networks as central image-enhancement blocks.
 This is in line with recent state-of-the-art architectures, which also make use of various advanced sub-networks, e.g., see~\cite{kno+20b,muc+20,hsqdsr19,ramzi20,sriram20}.
 Somewhat surprisingly, it turned out to be beneficial that the same $\UNet$ is used in all four iterations (weight sharing), cf.~\cite{amj18,hsqdsr19}.

 \item We have observed that it is crucial to \textbf{initialize} the $\UNet$-parameters $\tilde{\NNparam}$ by the post-processing weights from Step 2.
 This does not only increase the speed of convergence, but it also significantly improves the final accuracy (see Fig.~\ref{fig:train}).
 In other words, our results show that the initialization of the $\UNet$-block as a post-processing unit makes it possible to find better local minima.
 To the best of our knowledge, such an effect has not been reported in the literature yet.
 We emphasize that this initialization strategy is enabled by making use of a powerful enough post-processing sub-network.

 \item  Our \textbf{data-consistency} layer is inspired by a gradient step on the loss $\x \mapsto \tfrac{\dcparam_k}{2} \norm{\AFan \x - \y}_2^2$, which would result in the update $\x \mapsto \x - \dcparam_k \cdot \AFan^\T (\AFan \x - \y)$.
 We depart from this scheme by replacing the unfiltered backprojection $\AFan^\T$ by its filtered counterpart $\FBP$.
 This modification leads to significantly improved results for two reasons: (a) it counteracts the fact that the unfiltered backprojection is smoothing; (b) it produces images with pixel values at the right scale.
 Therefore, we interpret the resulting $\ItNet$  as an industry-like iterative CT-algorithm (e.g., see~\cite{willemink19}), rather than a neurally-augmented convex optimization scheme.
\end{enumerate}

In our experiments, we witnessed only minor effects by computing more than four iterations in~\eqref{eq:itnet}.
However, the accuracy was improved by the following \textbf{post-training} strategy:
First, the $\ItNet$ is extended by one more iteration:
\begin{align}
 \label{eq:itnet-5}
 \ItNett[\NNparam] \colon \R^m \to \R^N, \ \y \mapsto \left[ \bigcirc_{k=1}^5  \left( \DC_{\dcparam_k,\y} \circ \U[\tilde{\NNparam}_k] \right) \circ \FBP \right](\y),
\end{align}
where $\tilde{\NNparam}_k$ is initialized by the optimized weights of~\eqref{eq:itnet} for $k=1,\dots,5$.
Then, $\ItNett$ is fine-tuned by keeping the weights $\tilde{\NNparam}_1 = \tilde{\NNparam}_2 = \tilde{\NNparam}_3$ of the first three $\UNet$s fixed and training only the last two iterations (without weight sharing).
The obtained improvements indicate that there is a trade-off between increasing the model capacity by more iterations and the difficulty of optimizing the resulting network.
The systematic study of such iterative training strategies is left to future research.

The initialization and training of the \textbf{weights} $\dcparam_k$ has a considerable impact on the accuracy of $\ItNet$ and $\ItNett$.
We have found that $\DCparam =[\dcparam_1,\dcparam_2,\dcparam_3,\dcparam_4]$ typically converges to values of the form $\{\dcparam_1 <\dcparam_2<\dcparam_3\gg\dcparam_4\}$ after sufficiently many training epochs of $\ItNet$.
For an additional speed-up of the training, we use the initialization $\DCparam = [1.1, 1.3, 1.4, 0.08]$, which was found by pre-training.
Similarly, $\ItNett$ is initialized with the final values of $\ItNet$ for $k=1,2,3$, together with $\dcparam_4=1.0$ and $\dcparam_5=0.1$.
We suspect that a systematic study of these scalar weights could lead to further performance gains and to a more regular training procedure.
In particular, it might be beneficial to decouple them from the training of the $\UNet$-weights.

\paragraph{Fine Tuning}

To improve the overall performance of our networks, we have additionally applied the following ``tricks'', which are ordered by their importance:

\begin{enumerate}[label={(\roman*)}]
 \item Due to statistical fluctuations, the networks typically exhibit slightly different reconstruction errors, despite using the same training pipeline.
 For the computation of our final reconstructions, we therefore \emph{ensemble} 10 networks, each trained on a different split of the training set.

 \item  Due to the training with small batch sizes, we replace batch normalization of the $\UNet$-architecture by \emph{group normalization}~\cite{wu18}.

 \item We equip the $\UNet$-architecture with a few \emph{memory channels}, i.e., one actually has that $\U[\NNparam] \colon \R^N \times (\R^N)^{c_{\textrm{mem}}} \to \R^N \times (\R^N)^{c_{\textrm{mem}}}$ (cf.~\cite{pw17,ao18}). While the original image-enhancement channel is not altered, the output of the additional channels is propagated through $\ItNet$, playing the role of a hidden state (in the spirit of recurrent NNs). For our experiments, we have selected $c_{\textrm{mem}} = 5$.

 \item It was beneficial to restart occasionally the training of the networks, e.g., see Fig.~\ref{fig:train}.

\end{enumerate}

\noindent The following modifications did not lead to a gain in performance and were omitted:

\begin{enumerate}[label={(\roman*)}]

 \item Improving the $\FBP$ in Step 1 by making some of it components learnable (e.g., the filter), cf.~\cite{wurfl16}.
 Although this is advantageous for the reconstruction quality of the $\FBP$ itself, it leads to worse results for $\UNet$ and $\ItNet$.
 This suggests that a combination of model- and data-based methods benefits most from  precise and unaltered physical models.

 \item Adding additional convolutional-blocks in the measurement domain of $\ItNet$.

 \item Modifying the standard $\ell^2$-loss by incorporating the RMSE or the $\ell^1$-norm.

 \item Utilizing different optimizers such as RAdam, AdamW, SGD, or MADGRAD.

 \end{enumerate}

\section{Results}

\begin{figure}
 \centering
 \includegraphics[width=\textwidth]{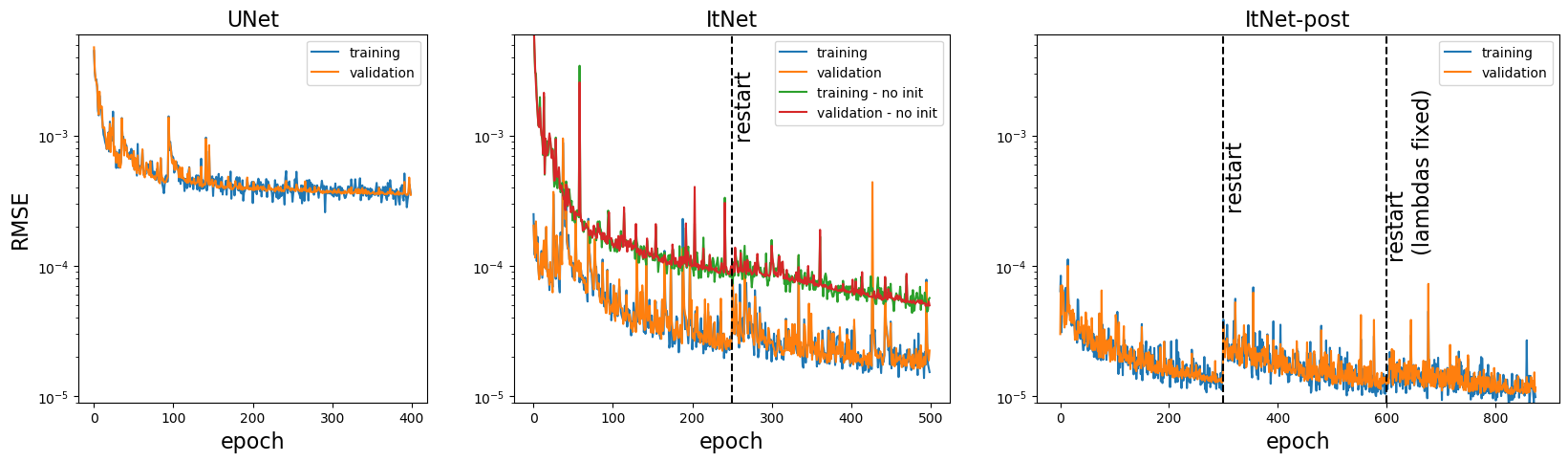}
 \caption{\textbf{Loss curves and network training.} The first two plots demonstrate that $\ItNet$ improves the RMSE by approximately an order of magnitude in comparison to a post-processing by $\UNet$. Furthermore, the gain of our $\UNet$-initialization strategy can be seen in the second graph. The last two plots illustrate the advantages of restarting and of the post-training strategy, respectively. Note that we display the RMSE on the training and validation sets instead of the actual $\ell^2$-losses, which behave similarly. }
 \label{fig:train}
\end{figure}

We conclude this report by briefly showing some of our results.
In terms of quantitative similarity measures, we restrict ourselves to reporting the RMSE, which is the main evaluation metric for the challenge.\footnote{Note that the RMSE is only reported for a subset of 125 images from the training set, which we have used for validation.
Hence, these values might differ from the actual results on the official challenge test set.
In particular, the values of the ensembled $\ItNett$ might (slightly) overfit, since we did not hold back separate validation images for evaluating the ensembling step. }
For comparison, we also consider a post-processing of the $\FBP$ by the more advanced \emph{Tiramisu-architecture}~\cite{bub+19,genzel20,jdvrb17}.
Furthermore, we have also trained the iterative \emph{learned primal-dual} (LPD) scheme~\cite{ao18} (slightly modified by replacing the unfiltered backprojection by the $\FBP$).

\begin{figure}
 \centering
 \begin{tabular}{cc}
  \includegraphics[width=0.5\textwidth]{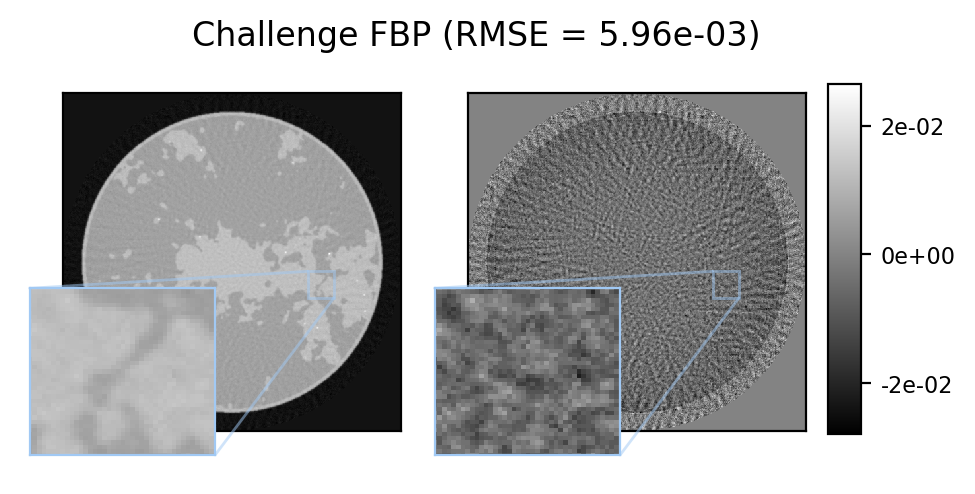} &
  \includegraphics[width=0.5\textwidth]{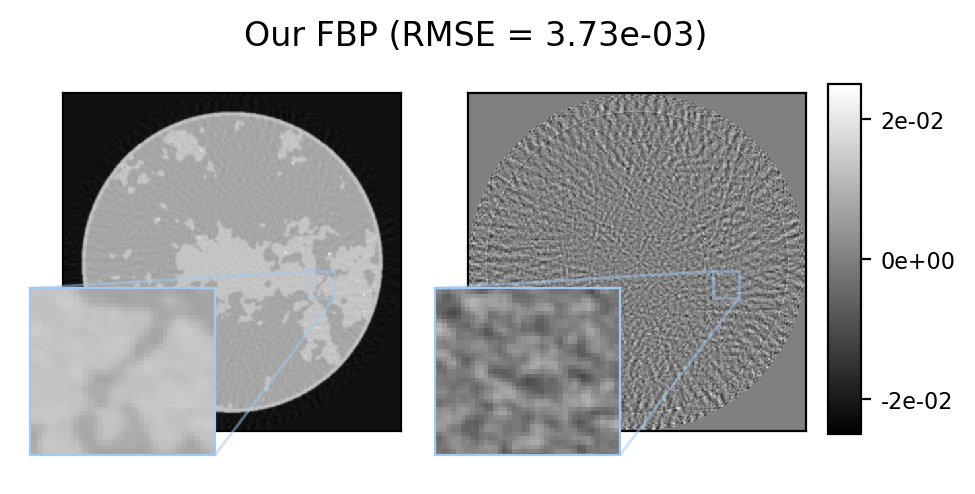} \\
  \includegraphics[width=0.5\textwidth]{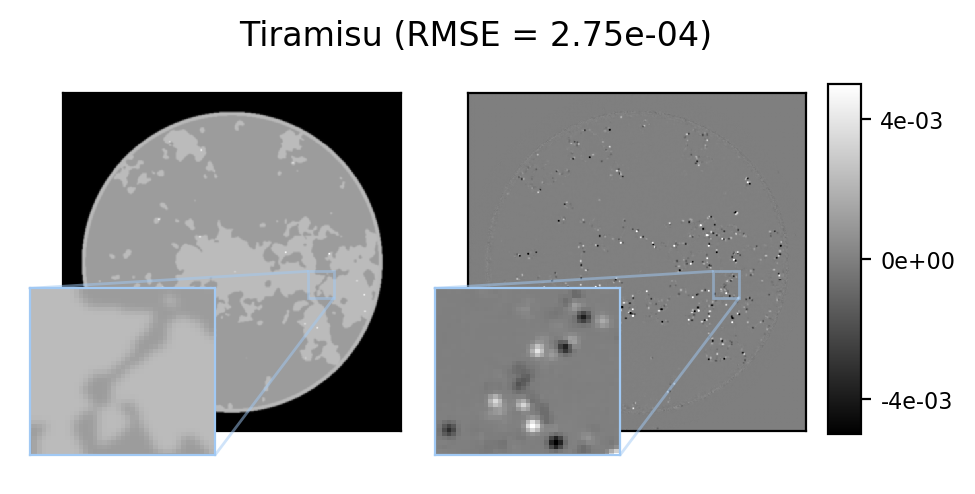} &
  \includegraphics[width=0.5\textwidth]{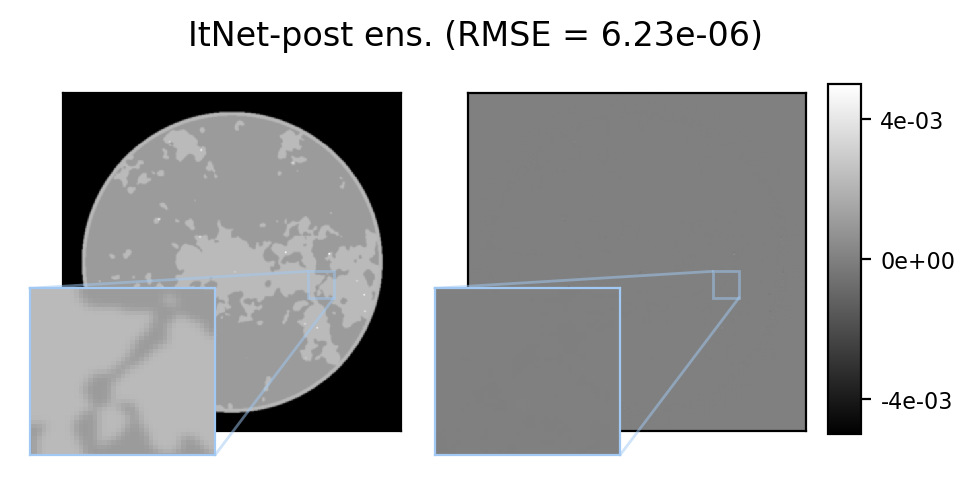}
 \end{tabular}
 \caption{\textbf{Reconstruction results.} We display reconstructions of a validation image. The first row compares the $\FBP$ provided by the challenge with our own (see Step 1). The second row compares a post-processing by Tiramisu with the (ensembled) $\ItNett$. The ground truth image is omitted, since it is visually indistinguishable from $\ItNett$.}
 \label{fig:recs}
\end{figure}

\begin{figure}
 \centering
  \includegraphics[width=\textwidth]{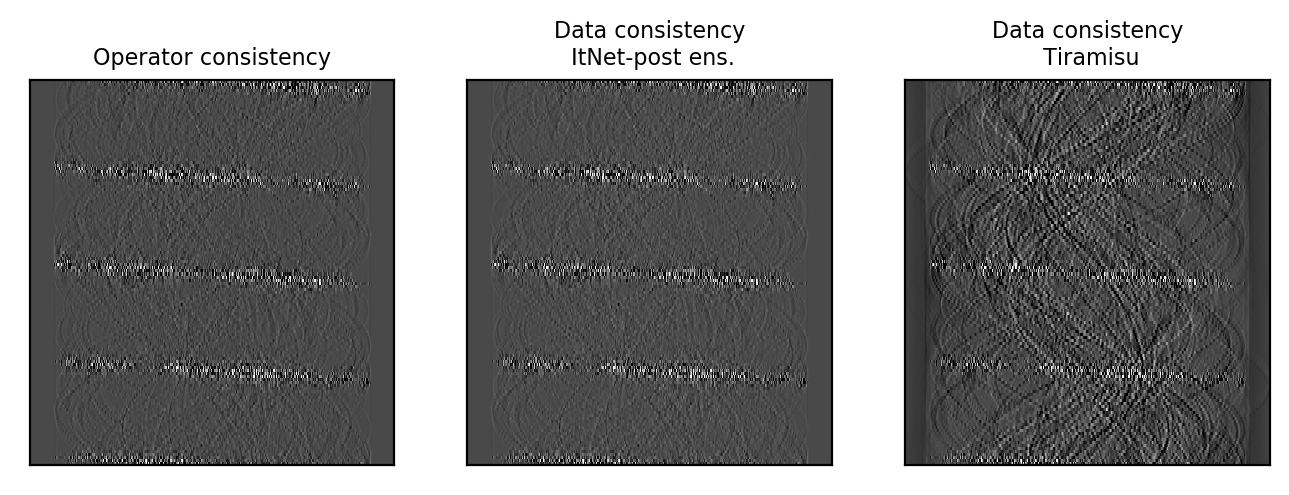}
 \caption{\textbf{Data-consistency.} The first image analyzes the accuracy of our forward model by displaying the error $\ygrtr - \AFan \xgrtr$ for a sinogram-image pair $(\ygrtr,\xgrtr)$ from the validation set. The visualization of $\ygrtr -\nobreak \AFan \cdot\nobreak \ItNett(\ygrtr)$ in the middle is visually nearly indistinguishable, which shows that $\ItNett$ inherits the inaccuracies from Step 1. Therefore, $\ItNett$ would allow for even better results if a more accurate forward model was available. It is also interesting to compare with $\ygrtr - \AFan \cdot \Tira(\ygrtr)$: The image on the right exhibits considerably larger errors, which reveals that the post-processing by $\Tira$ suffers from a lack of data-consistency. All images are shown in the same dynamical range.}
 \label{fig:sino}
\end{figure}

In Fig.~\ref{fig:train}, we first visualize the RMSE loss curves of our training pipeline, i.e., $\UNet \to \ItNet + \text{restart} \to \ItNett + 2\times\text{restart}$.
Furthermore, the average performance of all considered networks is reported in Table~\ref{tab:res}.
To give a visual impression as well, reconstructions of an image from the validation set can be found in Fig.~\ref{fig:recs}.
Finally, we analyze the aspect of data-consistency in Fig.~\ref{fig:sino}.
This figure suggests that the performance of $\ItNett$ could still be improved if the exact forward model was available.
This conclusion is also underpinned by the following observation:
Even a small improvement in the parameter identification of Step 1 resulted in a significantly more accurate $\ItNett$.

\begin{table}
\centering\footnotesize
\begin{tabular}{ccccccccc}
\toprule
& \multicolumn{2}{c}{Baselines} & \multicolumn{4}{c}{Our Network Variants} & \multicolumn{2}{c}{Comparison Networks}\\
\cmidrule(lr){2-3} \cmidrule(lr){4-7} \cmidrule(lr){8-9}
& FBP & Our FBP& $\UNet$ & $\ItNet$ & $\ItNett$ & $\ItNett$ (ens.) & $\Tira$ & LPD \\
\midrule
RMSE & 5.72e-3 & 3.40e-3 & 3.50e-4 & 1.64e-5 & 1.05e-5 & \textbf{6.42e-6} & 2.24e-4 & 1.24e-4 \\
\bottomrule
\end{tabular}
\caption{This table reports the average RMSE.}
\label{tab:res}
\end{table}

\renewcommand*{\bibfont}{\smaller}
\begin{refcontext}[sorting=nyt]
	\printbibliography[heading=bibintoc]
\end{refcontext}

\end{document}